\documentclass{article}

\usepackage{microtype}
\usepackage{graphicx}
\usepackage[export]{adjustbox}
\usepackage{subfigure}
\usepackage{booktabs} 

\usepackage{hyperref}

\usepackage[accepted]{icml2024}

\usepackage{amsmath}
\usepackage{amssymb}
\usepackage{mathtools}
\usepackage{amsthm}

\usepackage[capitalize,noabbrev]{cleveref}

\theoremstyle{plain}

\theoremstyle{definition}

\theoremstyle{remark}

\usepackage[textsize=tiny]{todonotes}

\usepackage{amsmath}
\usepackage{amssymb}
\usepackage{amsthm}
\usepackage{mathtools}
\usepackage{multirow}
\usepackage{xcolor}
\usepackage{bbold}

\usepackage{pifont} 
\usepackage{pgf}

\newcommand{\T}{Position: Explain to Question not to Justify
}
\icmltitlerunning{\T}

\begin{document}

\twocolumn[
\icmltitle{\T}
%




\begin{icmlauthorlist}
\icmlauthor{Przemyslaw Biecek}{uw,wut}
\icmlauthor{Wojciech Samek}{hhi,btu,bifold}
\end{icmlauthorlist}

\icmlaffiliation{uw}{MI2.AI, University of Warsaw, Poland}
\icmlaffiliation{wut}{MI2.AI, Warsaw University of Technology, Poland}
\icmlaffiliation{hhi}{Department of Artificial Intelligence, Fraunhofer Heinrich Hertz Institute, Germany}
\icmlaffiliation{btu}{Department of Electrical Engineering and Computer Science, Technical University of Berlin, Germany}
\icmlaffiliation{bifold}{BIFOLD - Berlin Institute for the Foundations of Learning and Data, Germany}

\icmlcorrespondingauthor{Przemyslaw Biecek}{przemyslaw.biecek@pw.edu.pl}

\icmlkeywords{trustworthy machine learning, interpretability, explanation}

\vskip 0.3in
]



\printAffiliationsAndNotice{}  

\begin{abstract}
Explainable Artificial Intelligence (XAI) is a young but very promising field of research. Unfortunately, the progress in this field is currently slowed down by divergent and incompatible goals. We separate various threads tangled within the area of XAI into two complementary cultures of human/value-oriented explanations (BLUE XAI) and model/validation-oriented explanations (RED XAI). This position paper argues that the area of RED XAI is currently under-explored, i.e., more methods for explainability are desperately needed to question models (e.g., extract knowledge from well-performing models as well as spotting and fixing bugs in faulty models), and the area of RED XAI hides great opportunities and potential for important research necessary to ensure the safety of AI systems. We conclude this paper by presenting promising challenges in this area.
\end{abstract}

\section{Fallacies behind the XAI crisis}

\begin{quote}
``Interpretability may be one of the most confused topics in all of machine learning, fraught with confusion and conflict. Read an interpretability paper selected at random and you’ll find representations (or insinuations) that the work is addressing “trust”, “insights”, “fairness”, “causality”. Then look at what the authors actually do and you’ll be hard-pressed to tie back the method to any of these underlying motivations. Half the papers produce a set of feature important scores.'' -- Zachary Lipton in \citet{goldblum2023perspectives}
\end{quote}

\begin{figure}[t!]
    \centering
    \includegraphics[width=0.45\textwidth]{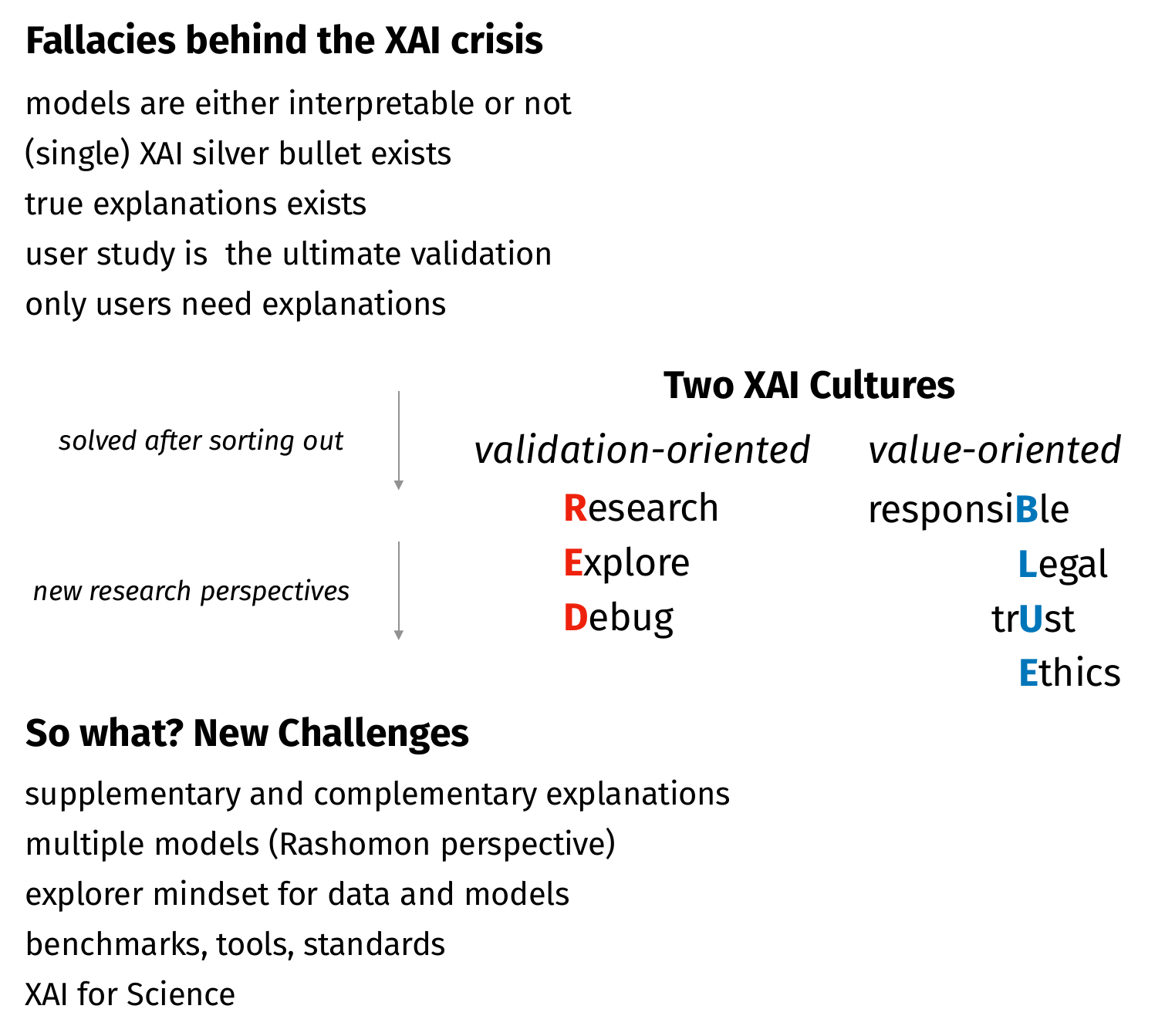}
    \caption{Summary of the main claims in this paper. The field of explainable machine learning (XAI) is currently experiencing a crisis of identity. Misconceptions about the role and goals of the XAI method are partly responsible for this crisis, and in the first chapter, we discuss the most prominent misconceptions. Having diagnosed the problems, we find that there are two communities of researchers working in the XAI field having different goals. In the second chapter, we identify these cultures and disentangle the main goals and motivations behind each. In addition, we find that the current narrative in survey papers is dominated by a discussion of the goals of the BLUE XAI culture. However, it is the RED XAI culture that has very promising research challenges ahead. A key conclusion of this diagnosis is to identify new challenges and research areas that concern RED XAI culture. These are presented in the third part of this work.}
    \label{fig:fig1}
\end{figure}

The growing number of developments that incorporate artificial intelligence and machine learning inevitably leads to an increasing number of examples of AI models being flawed \cite{AIIncident2023}. Failures in the functioning of these models have become the driving fuel for eXplainable Artificial Intelligence (XAI) research \cite{darpa-xai}. One would think that the discussion of the AI act \cite{eu03} or Executive order on AI safety \cite{Executive-Order} would follow the flourishing of new XAI techniques. But this is not the case. Instead, the field of XAI is experiencing an confusion in the flurry of expectations, troubled by vague and diverging goals.

Expectations for XAI methods are very high and broad. Models that everyone can trust, everyone will want, and be able to understand, that will perform better than humans guided by human values and judgment? Universal transparency that is a panacea for imperfect data and algorithms? In this paper, we challenge this way of thinking, demonstrating that it is unrealistic and harmful. 

An example of the harmful effects of an excessive focus on simplicity of explanations can be seen in the frequent expectation of reviewers that new XAI methods should be tested using user studies on a wide range of users. The focus on validation through user studies can make it difficult to publish advanced explainability techniques that are aimed exclusively at advanced AI developers \cite{AndIF22, PahArXiv22}.

Even popular explanation techniques such as SHAP or LIME may be very difficult to understand without some level of technical knowledge. So, instead of relying on explanations served to end users, it is wiser to design explanations to empower model developers. If we want to make progress toward safe AI, then we need new techniques for exploring and debugging models to be used by AI professionals.
The assumption that every user wants to decipher explanations may not be correct and at the same time may obstruct efforts to develop advanced and complex model exploration techniques that will be useful to advanced users.
Further progress in XAI research is essential, but it also needs the right motivations. 

Before proposing a solution, let us deconstruct five fallacies that are often wrongly imputed to the entire field of explainable artificial intelligence.

\textbf{Fallacy: Interpretability is a binary concept} and models can be divided into interpretable vs.\ black boxes. Suggesting that models can be divided into transparent and black-box models is false and harmful. In the literature, models such as linear regression or decision trees are often referred to by the term transparent, thereby directing XAI techniques to these other complex models. Such a division is untrue because both linear and tree models can be very difficult to analyze if they are based on a large number of variables. As few as 10 parameters make a model difficult to analyze, and real-world applications are often based on hundreds or thousands of variables. Such partitioning is also harmful because it creates the illusion that transparent models are natively correct and free of bias. And this is not true. 
In our opinion, XAI techniques should be viewed as a magnification lens, which can provide helpful insights, both for ``transparent'' and ``black-box'' models. 

\textbf{Fallacy: Single XAI silver bullet exists} and we just need to find this single best XAI method. 
Many papers on explainability are structured in such a way that first, they present desirable properties for explanations, and later they present a method that has these properties. For example, in a paper on SHAP, these properties are Local accuracy, Missingness, and Consistency, while in the paper about LIME, these properties are Local fidelity, Model-agnostic, using a sparse number of interpretable features. 
While it is certainly useful to investigate the theoretical properties of explanation methods, one should be careful to not create the false and harmful impression that model explanations will come down to using one single right method. Since different methods often explain (slightly) different things (e.g., relevance, positive relevance, relative relevance, sensitivity, interaction) and make different assumptions about the model and the data distribution, there is no almighty explanation method (of course one explanation method can be more appropriate than another in a specific context).
In our opinion, it is perfectly valid to investigate different aspects of the model behavior by looking at it through different magnification lenses (aka different XAI methods). 

\textbf{Fallacy: The illusion of a ``true explanations''}, which need to be approximated by the XAI method. 
In certain applications, especially in medical image classification tasks, explanations are evaluated by whether they are consistent with the masks described by the domain expert. For example, if a model predicts tumors in x-ray images, it is assumed that the explanation for such classification should be a mask showing the tumor in the image. If for the classification issue such masks are available (the so-called Ground Truth for explanations) then the quality of the explanation can be described by the intersection of the generated explanation and the true explanation.
However, such a procedure has several weaknesses: (1) for some problems the ground truth cannot be constructed, (2) the mismatch between the explanation and the ground truth may not be due to a bad method of explanation, but, for example, to a bad model behavior which is interesting on itself \citep[see][]{saporta2022benchmarking,baniecki2023careful}, (3) some decisions may be based on several equally valid explanations. 
In our opinion, such Ground Truth-based thinking may be appropriate in certain highly-controlled settings \citep[e.g.,][]{ArrINF22}, but may be harmful and misleading in real-world scenarios, where we consider ``explaining'' to rather be an interactive process than a single one-shot explanation. 

\textbf{Fallacy: User-study is the ultimate validation of XAI methods} and such studies shall be required for papers introducing a new XAI method.
One of the unquestionable challenges in the construction of new explanation methods is the problem of their evaluation. As long as there are no developed evaluation protocols, reviewers often ask for user-studies so as to assess the explanations' usefulness. However, such validation is not always necessary or useful. If the XAI technique is used to validate the behavior of a model at the model development stage, the end user will not see such explanations and there is no need for them to be evaluated. A separate problem is the selection of the right users to test the explanation method. If the explanations are aimed at expert radiologists familiar with the technologies then testing inexperienced medical students may give a false picture.
In our opinion, a utilitarian perspective should be taken when evaluating XAI methods, i.e., a method is good if it provides a (measurable) benefit. Note that this benefit may be completely different for the lay user, the expert radiologist, and the AI developer.

\textbf{Fallacy: Explanations need to be aimed at users to increase their trust in explained models}. This is a fallacy similar to the previous one, assuming that all end users are willing and capable of reading and understanding the explanations. And at the same time, it leaves out other potential recipients of explanations, such as model developers.
In our opinion, there are different motivations to use explanation techniques, and fostering trust is just one of them.

In some situations, the properties described above are desirable. In others, they are harmful because they can slow down progress by setting the wrong priorities. Therefore, it is important to know in what context and purpose we are working with the explanations and to educate the user on these different properties and fallacies (instead of promoting an unreflected, out-of-the-box application of XAI methods). The next chapter is focused on the different goals and motivations of working with explanations.

\textbf{This position paper argues that besides being used for justification of model's decisions, new explanation methods are  needed to specifically question models, i.e., extract knowledge from well-performing models as well as spotting and fixing bugs in faulty models.}

\begin{table*}[th!]
    \centering
\caption{A comparison of two cultures of the work on explanation techniques. The table shows a representative typical perspective; in many applications, both perspectives are present to some extent.} \label{tab:twoxaiculuters}
    \begin{tabular}{p{4cm}|p{6cm}|p{5.7cm}}
     \toprule
         & Model-validation oriented  {\color{red}RED} XAI & Human-values oriented {\color{blue}BLUE} XAI \\ \midrule 
\textbf{Why} explanations are produced? & \ {\color{red}R}esearch on data, {\color{red}E}xplore models, {\color{red}D}ebug models & responsi{\color{blue}B}le models, {\color{blue}L}egal issues, tr{\color{blue}U}st in predictions, {\color{blue}E}thical issues  \\ 
\textbf{When} explanations are read and used? & Empower model developer, mostly during training & Empower user, mostly during model inference \\ 
\textbf{Who} is the direct audience of the explanations?  & Power user, Model developers, AI researchers  & Lay user, Customer, Patient \\
\textbf{What} are desired characteristics of explanations & Faithful to model and data, Actionable & Simple and easy to understand \\
          \bottomrule
    \end{tabular}
\end{table*} 

\section{Two XAI cultures}

We identify two cultures of thinking about explainability -- one centered on human values such as fairness, ethics, trust and the other centered on model validation, exploration, and research on model and data. We describe the similarities and differences between these cultures and argue that relatively little attention has been paid to the model validation centered culture.
This undifferentiated view potentially pushes us away from working on exciting and promising research directions. If we move aside expectations typical of human values-centered explanations, we find that many promising research directions open up for model validation-centered explanations. We present examples of such challenges in the third part of this paper.

In 2001, Leo Breiman published a seminal article, ``Statistical Modeling: The Two Cultures'' \cite{breiman2001statistical} in which he diagnosed the growing gap between two cultures of researchers working on statistical models, a culture focused on understanding the data and an algorithmic culture focused on predictive performance.
Breiman's motivation was to encourage statisticians to work on interesting challenges that are overlooked in the data understanding culture, but at the same time which play a big practical role.
The article was like spring rain, both waking up from the lethargy of working on unrealistic theory and providing a reflection of many interesting research developments.

A similar growing division is also evident among those working on explainable artificial intelligence (XAI). Scientific papers often migrate towards focusing on values such as fairness, ethics, and trust. While in applications, the issue of model performance is often of primary interest, where XAI can also help a lot. 
Moreover, without a better understanding of the different goals facing these cultures, further progress in any of them will be significantly hindered.

In this paper, we identify two cultures of thinking about explanations, which we will hereafter refer to as BLUE XAI and RED XAI\footnote{The names RED and BLUE are not only acronyms for the goals of the two communities' explanations (see Table 1), they are also a reference to the Red team and Blue team roles defined in IT security systems. The term Blue team is used to describe people with a mindset of defending the IT system, while the Red team is the mindset of those trying to adversarially find and exploit vulnerabilities in the IT system in order to fix such issues. We decided to apply this nomenclature to XAI cultures as well. In a natural and intuitive way, the Blue team is focused on security and human values and the Red team is focused on fully understanding the behavior of the model as well as its vulnerabilities.}. These two cultures differ significantly in their expectations, methods and approaches. A summary of the two cultures is presented in Table \ref{tab:twoxaiculuters}.

At the very least, work on explainability is carried out with at least two goals:
\textit{High trust towards the human-aligned operation of the model}. If the model's performance is predictable then it will be used more often or one can learn something from the model. Determining whether the factors influencing the model prediction are aligned with acceptable norms in the specific application area.
\textit{High model accuracy}. An important goal in modeling is high performance, not only in training or test data but also in the future, after model deployment, under adversarial scenarios, or as a result of data drift.

There are two different approaches toward these goals:

\textbf{BLUE XAI: Human Values Oriented Explanations}  primarily designed for final users of a model. 
In this case, the intended recipient of the explanations is the user of the AI system, such as a bank customer in the case of a credit risk scoring system, or a patient in the case of AI systems used in healthcare. In many situations, such users have low level of technical background about AI systems, so when aiming an exception to the user, the key theme becomes simplicity and comprehensibility. For such lay users, one of the key questions is whether the user believes or trusts the AI system because otherwise, the user will not accept the system's prediction.

One way to think about explanations is from the perspective of a citizen who receives a decision that he/she doesn't understand, disagrees with, or is curious about, and asks 'where did this decision come from?' Whether it involves credit, access to medical services, or restaurant recommendations is secondary. The important thing is that the person asking 'Why' is the user of the AI system. 

This perspective places emphasis on the model's decision received by a particular individual and assumes that explainability will help him or her accept or challenge that decision.

Analysis of a model in this culture is oriented toward user rights (e.g. famous ``right to explanations''). Discussion of the need for and desirability of explanations revolves around such concepts as trust, right to explanation, causality,  fair and ethical decision-making.  An example paper that has resonated and illustrates well the perspective of this culture is ``The Mythos of Model Interpretability'' \cite{Lipton2018}.



\textbf{RED XAI: Model Validation Oriented Explanations} primarily designed for model developers. In this case, the intended recipient of the explanations is an AI expert who trains, audits, debugs, and verifies or improves the AI system.
Depending on the application, such a power user may have access to internal model parameters and training data, but may also operate on a ready-trained model. What distinguishes it is a high level of technical knowledge. It can be assumed that such audiences have strong analytical skills, so the explanations presented to them can be complex and delve into the details of how models work. When targeting explanations for a developer, the depth of exploration and the ability to understand how the model works becomes crucial. We can talk about whether the explanations faithfully reflect how the model works.

Another perspective on thinking about explanations is through the eyes of an engineer building or testing a model, who wants the model to work in every, or almost every, case. Explanations can help diagnose and perhaps fix errors in model behavior.
This perspective emphasizes the model that should work well for all subsets of data \cite{AndIF22}. 

Analysis of a model in this culture is oriented toward new ideas on how to detect a weakness in the model that can be corrected and thus lead to model improvement. 
Typical questions asked by this culture are if explanations are fidel to the model, give a global perspective, can be used to debug models, and can be used to improve models. A very influential paper that gives this perspective is the ``Why should I trust you'' paper \cite{ribeiro2016why} with a memorable picture of an incorrectly classified Husky dog.



\textbf{Prevalence of RED XAI and BLUE XAI perspectives in recent surveys on XAI}. In this section, we present highly cited review papers on XAI published in the last year. These papers were identified based on the Google Scholar database.

It turns out that the BLUE XAI perspective is very popular and many reviews are focusing on it. One of the more frequently repeated terms in 2023 is trustworthiness.  \citet{chamola_review_2023} discuss the key aspects and principles of trustworthy AI systems, such as Human Agency, Human Oversight, Transparency, Diversity and Fairness, and Accountability. In a similar style, the review by \citet{ali_explainable_2023} is conducted. The main motive behind this paper is that the model should be explainable to any user. Among the desirable characteristics that models should meet we find fairness, user satisfaction, and interpretability issues. In addition, a list of features that are necessary for explanations to be human-friendly is presented, such as Privacy, Causality, Factuality, Sociological
Understandability and Satisfaction.

Another group of review papers pays attention to the human aspect of XAI. \citet{severes_human_2023} discuss the issue of trust with a non-technical audience in mind. 
The same perspective is shared by  \citet{embarak_decoding_2023}, where the main threads revolve around user rights and human alignment.
An example of such a non-technical audience is the applications of XAI in education discussed in the review by \citet{farrow_possibilities_2023}. Among the challenges cited are the typically BLUE XAI themes of Confidentiality, Complexity, Unreasonableness and Injustice. The problem of the skill level of people reading explanations and the cognitive load of explanations is also discussed in the review by \citet{saeed_explainable_2023}.


\begin{figure*}[t!h]
    \centering
\includegraphics[width=0.87\textwidth,valign=t]{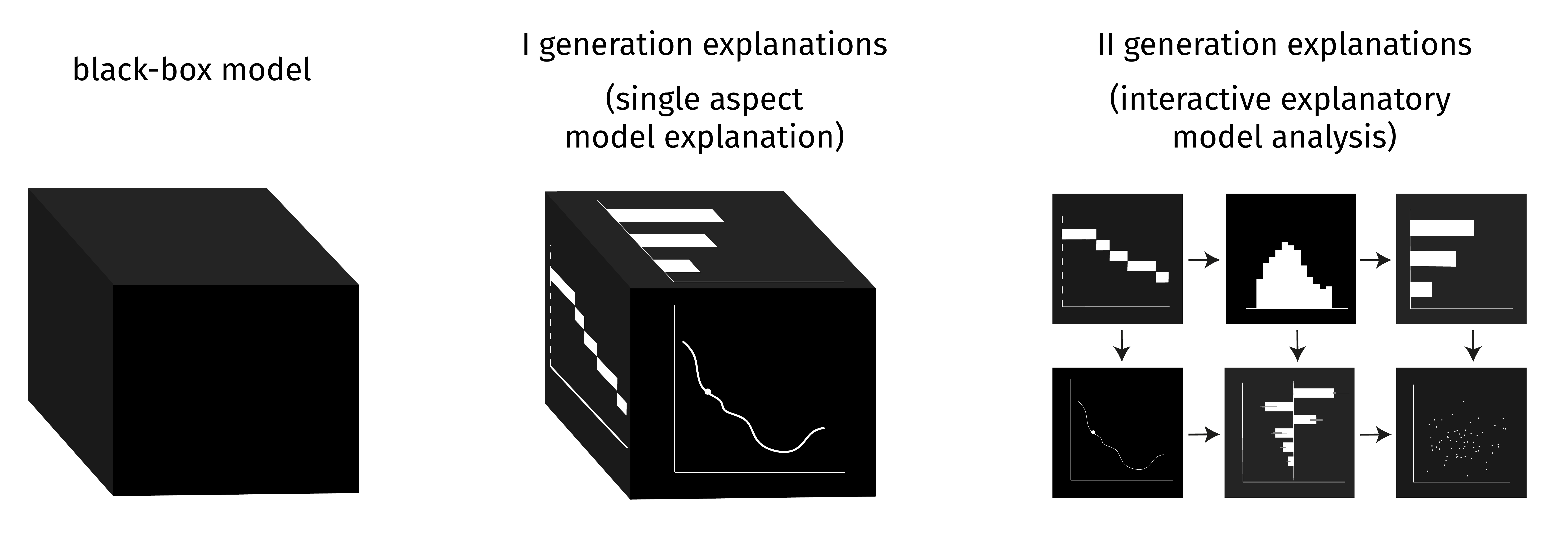} 
    \caption{Schematic diagram of multifaceted explanations. Many explanation techniques present one aspect of how models work. Multifaceted explanations show several complementary aspects simultaneously. Figure adapted from \cite{baniecki2023grammar}}
    \label{fig:iema_01}
\end{figure*}

A major challenge in the XAI area is how to evaluate proposed methods. The more difficult or fuzzy is the evaluation criteria, the harder it is to make progress in the area. A recent and comprehensive review paper devoted to analyzing metrics for XAI is published by \citet{sisk_analyzing_2022}. It makes significant reference to non-technical audiences, and with them, criteria such as Human Values or Inequality appear. Other review papers on evaluation (e.g., by \citet{speith_new_2023}) mention both RED XAI (e.g., debugging) and BLUE XAI (e.g., confidence, trust, usefulness, usability) measures for evaluating XAI threads.
Also, the review by \citet{reddy_explainable_2023} points out that there are no universal standards for evaluating XAI methods.


RED XAI perspective is less present in the current literature, but it also occurs. 
\citet{borys_explainable_2023} discuss various explanation techniques designed to analyze vision models in medical applications. 
Several complementary explanation methods are discussed for the diversification of XAI, allowing for the systematic uncovering of model flaws, data outliers, and biases.
Also \citet{WEBER2023154} focused strongly on debugging of AI models using XAI techniques. However, such reviews are still decidedly few in number for the potential of this direction.

\textbf{Why two cultures and not five or ten?} Many of the review papers discussed in the second section identified a number of XAI stakeholders. While such fragmentation often allows us to capture all or almost all potential uses of XAI, it does not create a mental mindset that is easy to remember and then apply. Therefore, in this paper, we have intentionally reduced the multitude of XAI applications by reducing them to two archetypal cultures. We are fully aware that there will be applications that do not fit into the two sketched poles. This, however, is fine. Untangling the research threads in the XAI community into the two described cultures clarifies thinking about new research directions, and this clarification is what we aimed for.

\section{So what? New challenges}

What follows from the diagnosis of the existence of two different cultures in the XAI community? Our position is that RED XAI brings with it many new open research questions that can be addressed independently of the challenges of BLUE XAI. In this section, we present selected challenges. However, they should not be treated as a closed catalog. Instead of this, we hope that these examples will provoke researchers to propose further interesting problems that can help advanced users in their work on improving predictive models or understanding how the models in question work.

\textbf{Challenge: construction of complementary explanations}. There is no single best technique for explanations and we need a multi-faced approach to model analysis.

Multi-faceted explanations take inspiration from the story,  ``Blind men and an elephant''. In this story, several blind people touch different parts of an elephant and have different ideas about what they are touching. Whether it is a rope or a trunk or a spear or something else. Each of these perspectives depicts a part of the elephant, but only by noticing all of them, one can correctly guess what animal they are dealing with. This inspiration can be transferred to explain complex models.


Explanation techniques present different, often complementary perspectives on the model's operation. A significant challenge is the juxtaposition and exploration of sequences of such perspectives using static or dynamic juxtapositions. Very complex models cannot be accurately described by a single approximation. To better understand how such a model works, several complementary explanations should be put together. See an example presented in Figure \ref{fig:iema_01}.

One of the earliest papers arguing that a single explanation is not enough to cover all aspects of complex models is \cite{sokol-interactive-customizable-explanations}. This paper describes the concepts of a dialogue system that uses different explanations in discussion as different arguments in the explication of predictive models.
A similar concept was implemented in by \citet{kuzba-what-ask-ml} based on the DialogFlow framework. The dialogue system presented in this work allows showing different explanations depending on the user's current intentions.

An interesting solution demonstrating the potential of interactive model exploration is the \textit{The What-If Tool} \cite{wexlerwhatif2019}. It allows quick summarization of model behavior by interactively selecting variables. Combined with an intuitive interface, this creates many possibilities.
The use of interactive tools for exploring models coded with TensorBoard is implemented in the \textit{explAIner} \cite{spinner-explainer}. Similarly, \textit{i-Algebra} implements a different approach to intertextual explanations based on drill-down concepts \cite{i-algebra}. \cite{teso2019explanatory} go one step further and propose an explanatory interactive machine learning approach.
Yet another example of how to juxtapose different explanations for vision models with each other is the \textit{iNNvestigate} package \cite{iNNvestigate2019}. 

The ever-increasing number of new model architectures and techniques for showing the various aspects of how these models work opens up a wide field of research into how to put these various aspects together to complement each other and enhance the capabilities of the person exploring the model.

\textbf{Challenge: need for benchmarks, tools, standards}. RED XAI is more engineering-focused, which also means it can benefit more from automating and streamlining the way models are explored. 

In questions related to model debugging, it is possible to create benchmarks to track the progress of the discipline. 
An example of such a benchmark in XAI is the FICO challenge \cite{fico2018}. Even if the part related to explainability was not measured in a simple algorithmic way, the competition itself inspired many interesting results.
In other disciplines, there are contests that sometimes last years, like Netflix prize \cite{bennett2007netflix} or CASP challenge \cite{moult_largescale_1995}, that have become fuel for the development of new methods.

The model exploration tools used in RED XAI can be used for a variety of applications. More versatile tools and processes on how to use these tools are needed to facilitate this exploration.
An example of a toolbox used in ML is sklearn library \cite{scikit-learn} that is now an industry standard. Similar tools are needed for RED XAI community, and they may be based on existing libraries like iNNvestigate for analysis of computer vision models \cite{iNNvestigate2019}, DALEX \cite{ema2021} to analyze models for tabular data, Quantus for evaluation of explanations \cite{Hedstrm2022QuantusAE} or others.

\textbf{Challenge: for RED XAI we need to take an explorer mindset}. Exploration focuses on the fact that sometimes we do not know what kind of signals and relationships we will find (so-called unknown unknowns) both by analyzing the models and by analyzing the data through the lens of the model.

The purpose of the benchmarks is not to identify a single explanatory method that will be the most effective. In the goals set for RED XAI, the wider the range of explanatory techniques, the greater the progress in the field. Benchmarks thus indicate the potential for creating new techniques. But one should look for complementary techniques or different ways to use certain techniques.

Not only is one explanation not enough, but no closed catalog of explanations will suffice for an in-depth analysis of a model and exploration of its behaviors. 
The process of model analysis needs to be automated by creating more ``checks'' and by developing techniques which systematically analyse and integrate the different views on the model, e.g., by meaningfully clustering the explanations \cite{DreArXiV23}, so that the ``essense'' of the model behavior can be shown to the user for a quick validation. However, even with the availability of such advanced tools, model exploration will always be a creative task requiring knowledge, skills, and imagination.

We need such a new mindset in model development, one that takes into account the analysis of not only the performance itself but also the (systematic) analysis of the model's behavior.

XAI techniques can be very useful here and we believe that there is a need for a new program in Machine Learning education that is more holistic and XAI/debugging centric.

\begin{figure}
    \centering
    \includegraphics[width=0.49\textwidth]{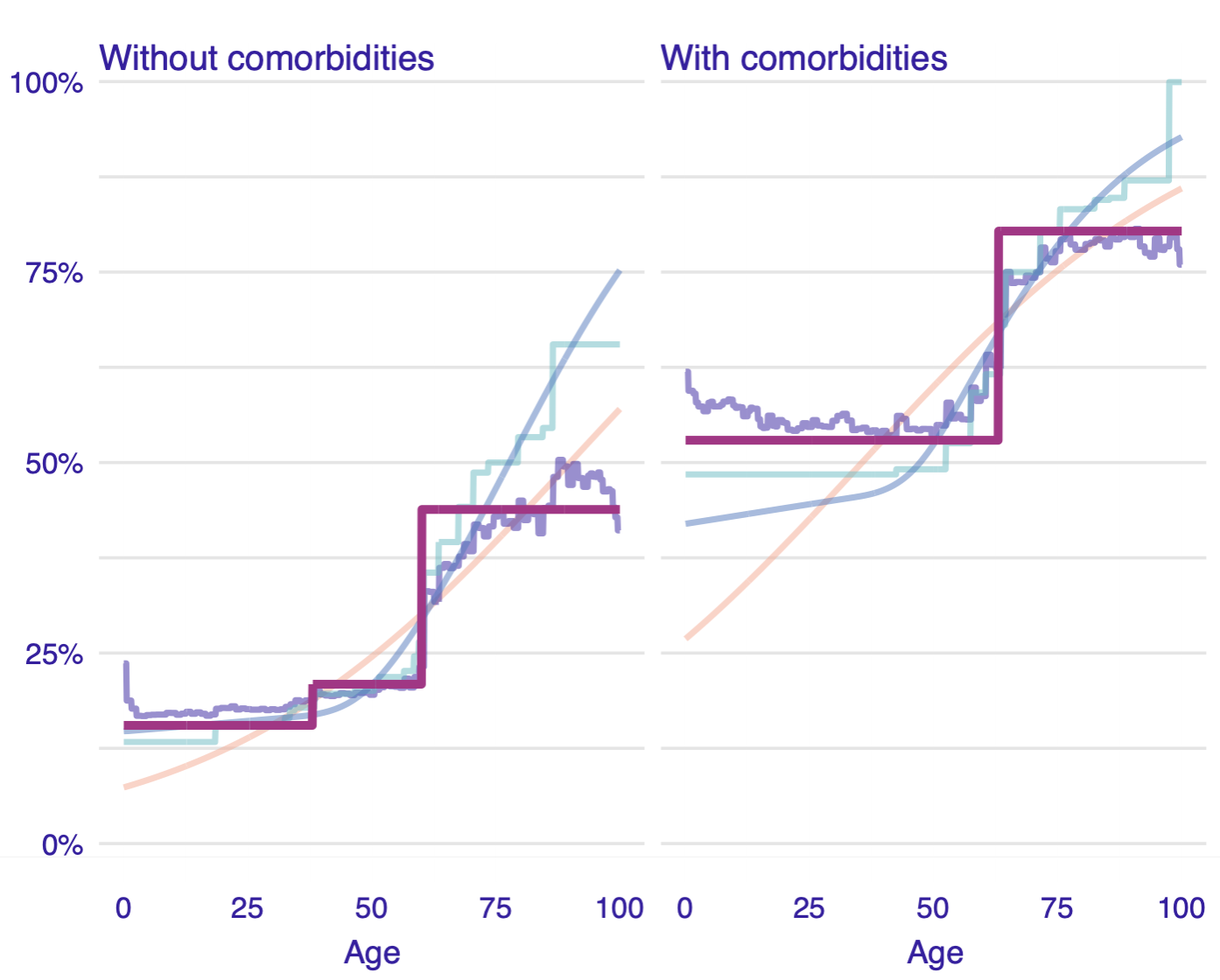}
    \caption{Partial Dependence profiles for five different predictive models (linear model, decision tree, random forest, boosting model) for Covid-19 data. Each color indicates a different model fitted to the same data. The profiles show the conditional response of a model predicting mortality conditional on two variables: age and the presence of concomitant diseases. Figure adapted from \cite{RML}}
    \label{fig:rashomon_01}
\end{figure}

\textbf{Challenge: multiple models and Rashomon explanations}. In order to understand the strengths or weaknesses of one model, it is sometimes necessary to contrast it with another model or models. To this goal it is beneficial to explore a Rashomon set of good models, that is, a set of models with very similar performance but  different behaviors that arrive at it. The purpose of such analysis may be to determine in which situations these models behave similarly and in which they act differently. An example of such a juxtaposition could be a comparison of two models, such as the champion-challenger style, the model currently used (champion) versus a newly trained model (challenger).

The concept of the Rashomon set is introduced by \citet{breiman2001statistical}. By using this name he relates to an Akira Kurosawa film called Rashomon, which depicted the story of a murder presented by four witnesses. Although each story describes the same event, it does so in different ways. Breiman used the name to describe a set of models that have the same performance but describe the data in different ways. The Rashomon set concept stands in opposition to the procedure in which data leads to a single model with the highest fit.
Since then, the concept of using a set of similar models has been increasingly researched, though still less than it deserves.

The most known application of a collection of different but equally good models is the bagging technique with its most famous representative - the random forest model \cite{randomForest}. Another application is the Rashomon ratio proposed by \citet{Semenova} examines the Rashomon set to determine the level of complexity of the problem, and thus the chance of finding a simple but equally good model. 

Characterization of the Rashomon set is listed as one of the main challenges of interpretable artificial intelligence \cite{challenges}. An example of the application of the Rashomon set is Model Class Reliance (MCR), which are intervals describing the importance of a specific variable for models from the Rashomon set. The concept of such intervals was presented in a model agnostic approach by \citet{rudin} and \citet{dong}. This concept for random forests was further explored in the work of \citet{RFMCR} and for decision trees in the work of \citet{xin}.

Exploring the Rashomon set using XAI techniques opens up many interesting challenges. If all the models in the Rashomon set are good, then we can select models from it based on other criteria, such as robustness, compatibility with domain knowledge, and reduction of possible discrimination. If we use the models not to determine predictions, but for data mining, then analyzing the Rashomon set will also allow us to understand whether the data has one unambiguous description or several divergent descriptions. An example of the application of the popular XAI technique, Partial Dependence Profiles \cite{Friedman00greedyfunction}, to the visualization of selected models from the Rashomon set, is presented in Figure \ref{fig:rashomon_01}.

\textbf{Challenge: XAI for Science}. Explanations may be used to generate and validate research hypotheses. 
A noble example of this is the ``Discovery of a structural class of antibiotics with explainable deep learning'' paper
\cite{wong_discovery_2023} which showed how model explanation and exploration tools allow us to better understand the data and domain, which can lead to interesting scientific discoveries. Or
``Acquisition of chess knowledge in AlphaZero''
\cite{mcgrath_acquisition_2022} Where analysis of a chess-playing model of super-human performance makes it possible to answer or verify many hypotheses related to the game of chess, such as the potential for different responses to different openings. While the use of explanations for knowledge discovery is a topic in itself and is already being addressed in different works, we believe that it fits well into the RED XAI perspective, with all the aspects discussed above. We encourage researchers to leave aside the user-oriented and silver bullet thinking and start using complementary methods and developing new tailored XAI tools (e.g., explanation techniques tailored to the specific data modality used, meta analysis tools summarizing and integrating the explanations), which help them to identify interesting relationships in the data and explore the unknown unkowns.

\section{Discussion}

In this section, we include topics pointed out during the review process. Readers are encouraged to expand or add more topics to the two XAI cultures. 

\begin{figure*}[h!]
    \centering
\includegraphics[width=0.85\textwidth,valign=t]{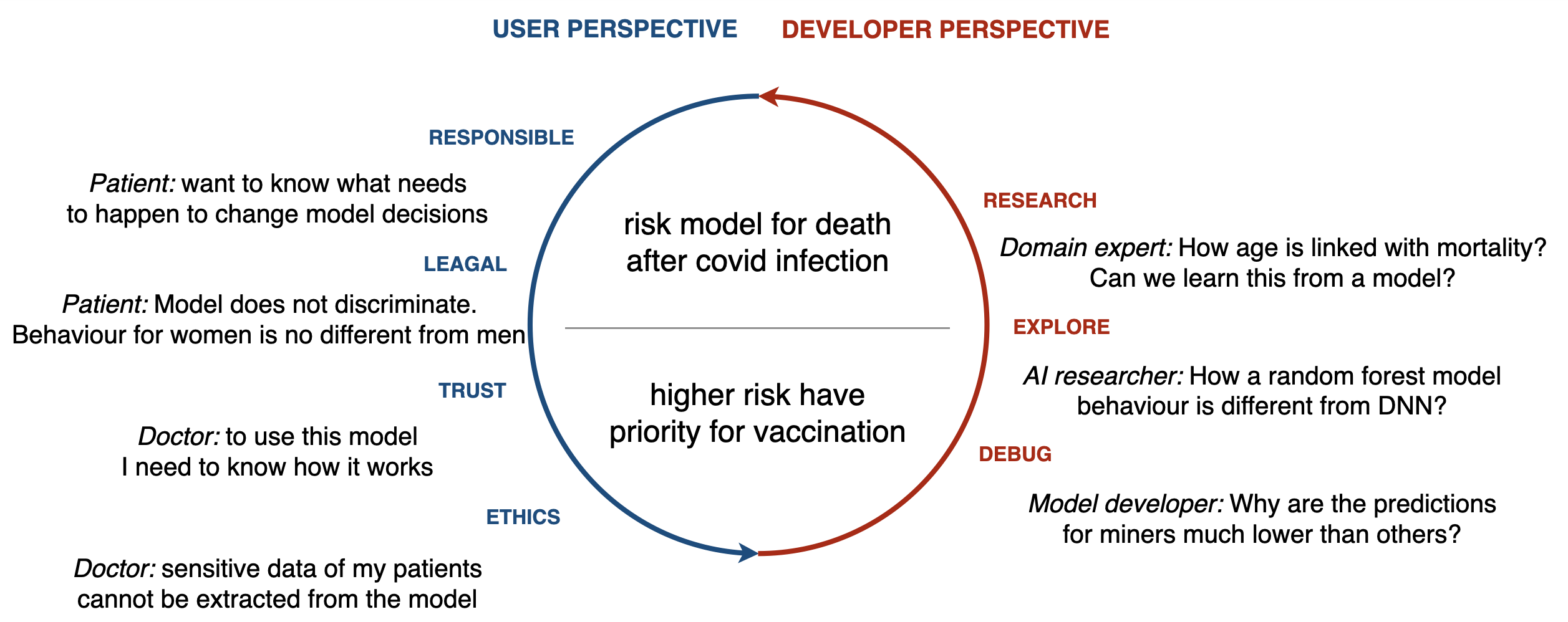} 
    \caption{Illustration of questions and needs raised by stakeholders interested in Blue XAI and Red XAI techniques. Questions are framed using the example of a hypothetical covid death risk assessment model used to prioritize access to vaccination. For each of the seven example areas, a sample question is presented that can be answered using XAI techniques. }
    \label{fig:fig4}
\end{figure*}

\textbf{Start explaining models for high-stakes decisions}. 
The need for explainability is often questioned and criticized. One of the most well-known arguments against explaining complex models are presented in the paper ``Stop explaining black box machine learning models for high stakes decisions and use interpretable models instead''\cite{rudin_stop_2019}.
Note, however, that the criticism presented there relates to the objectives indicated with BLUE XAI. Under this general title, the author criticizes the use of models that may not work, but their behavior is masked or justified by XAI-washing.
The objectives in RED XAI are completely different and the RED XAI process itself can and should be used for any type of model, especially models used in high-stakes decisions. While it is certainly true that providing explanations to the lay user can lead to an unsubstantiated trust in the system \cite{weller2019transparency}, we strongly believe that an increased understanding of the functioning of the model and a RED team mentality are necessary ingredients to build truly reliable AI systems in the future.

\textbf{Frontier models} (aka foundation models), unlike classical predictive models, can be applied to a wide variety of tasks. Typically, new applications of such models were not considered in the process of training, and often they were not even known, and only the availability of the model made it possible to see new applications.
In such situations, model exploration and debugging techniques are even more essential. The more potential applications, the greater the potential safety problem.
The BLUE-XAI community may approach the topic of LLM/foundation models from a ``human alignment'' perspective. It expects models to respect human values and not discriminate, cheat, use toxic speech, or hate speech. Benchmarks are being created in these directions, verifying how truthful, ethical, and ``aligned'' the LLMs/foundation models are.
The RED-XAI community may approach this problem from a different angle. Understanding the model is needed to control the model. One example of such an application is an experiment with the editing property of latent representation \cite{meng2022locating}, to better understand how LLMs store facts. Such knowledge opens new opportunities, e.g. to force the model to forget the facts in question. Another approach is the study of grooking phenomenon \cite{nanda2023progress}, whose goal is to understand the more complex models through studying simpler/smaller models.

\textbf{Covid-19 use-case} shown in Figure \ref{fig:fig4} illustrates the differences and similarities between the Blue and Red XAI perspectives. The model for assessing the risk of death could be used in prioritizing access to certain medical interventions, such as access to vaccines. 
The Blue XAI part is focused on the perspective of the user, in the case of this model it could be the perspective of a patient or a doctor. The transparency of such a model is important to achieve a high level of trust from users. The Red XAI part is focused on what information can be extracted from the model (which is important in the case of a new disease about which little was known) and how the model can be more accurately tested and fixed.


\textbf{Purple XAI} The terminology of the Red Team and Blue Team is rooted in military and security. Over time, additional ‘colors’ have been introduced into the team catalog, and this may also be the case with XAI communities. For example, Purple XAI can denote methods that can be used by both the model developer and the end users (certainly the most popular SHAP methods fall into this group). Similarly, using the color scheme used in security, one might discuss Yellow XAI which is more focused on the creation of processes, frameworks, and benchmarks that evaluate XAI methods, without explicitly defining an audience.
However, no matter how many different subgroups of the XAI community can be distinguished, it is worthwhile for each of these communities to have clear objectives. 

\textbf{Secret sauce} The methods developed by Blue XAI are targeted at the end user and will therefore be publicly visible quickly. The nature of the methods used by Red XAI may be different, as the tools that lead to better models may be regarded as a competitive advantage and protected know-how that model-creating organizations may not want to share (whether for fear of model security or competition). It is therefore possible that some of the new Red XAI methods will be developed outside the public academic community, and this may also explain the lower visibility of such methods at present.

\section*{Acknowledgments}

PB acknowledges financial support from the SONATA BIS grant 2019/34/E/ST6/00052 funded by Polish National Science Centre (NCN). WS acknowledges financial support by the German Research Foundation (DFG) - Research Unit KI-FOR 5363 (project ID: 459422098).

\section*{Impact Statement}

This paper aims at advancing the field of Trustworthy Machine Learning. The work has no immediate social implications, although in the long run we believe that sorting out the threads related to explainability will lead to the development of better models.

\bibliography{references}
\bibliographystyle{icml2024}

\end{document}